\documentclass[11pt]{article}
\usepackage[margin=1in]{geometry}
\usepackage{booktabs}
\usepackage{amsmath}
\usepackage{graphicx}
\usepackage{hyperref}
\usepackage{xcolor}
\newcommand{\illus}{\textit{(illustrative, not separately measured)}}

\title{Decompose, Don't Expand:\\
Capability-Independent AI Systems by Construction}
\author{Tejaswi Gowda\\ Arizona State University\\ \texttt{tejaswi@asu.edu}}
\date{\today}

\begin{document}
\maketitle

\begin{abstract}
A recurring design choice in building AI systems is whether, when a task is too
hard for a model, to \emph{expand} the model (make it larger, better-prompted,
fine-tuned), or to \emph{decompose} the task (extract the parts that admit a
deterministic solution to a host, leaving the model only the irreducibly fuzzy
residue). We make three claims. First, that many tasks commonly delegated whole to
a model are in fact deterministic tasks with a small generative core. Second, and
our central empirical result, that decomposing such a task yields a measurable and
desirable property we call \emph{parity by construction}: on the extracted
sub-tasks the model-size performance curve becomes flat, because no model performs
them at all. We measure this across nine models spanning a 494M-parameter local
checkpoint to a frontier Opus-class model, a thirty-fold parameter range on the
local family alone: once the host resolves the context, operation selection scores
\emph{identically} for all nine models and selector resolution for eight of nine,
while the one task that does not decompose, labelling, remains sloped across the
same range. Because the task leaves the model, a stock checkpoint suffices: no
task-specific training and no fine-tuning. Third, that this is a general design
discipline, not a domain trick: we show the same move recurring across physics
setup, motion retargeting, and object-model editing, and in a Web-CLI tool family
(same author, so not an independent replication). We give a method for locating the
cut, the construction that makes a cut exist (a deterministic surface and a closed
grammar over it), and a way to make the residue safe (resolve the deterministic
context; verify the generative output), position the work against neuro-symbolic AI,
tool-augmented language models, and generate-and-verify approaches, and are explicit
about what is measured (one domain) versus illustrative (the rest).
\end{abstract}

\section{Introduction}

When a task is hard for a model, the default move in contemporary practice is to
\emph{expand} the model: a larger model~\cite{kaplan2020scaling,wei2022emergent}, a better prompt~\cite{wei2022cot}, task-specific
fine-tuning, more examples. Expansion makes a system's correctness a function of
model capability, with four consequences that are easy to overlook: the system is
always one model generation behind; it is non-deterministic (the same input can
yield different outputs); it is unverifiable (one cannot prove the output
correct); and it has no floor (a cheaper, older, or smaller model simply fails).
A concrete instance in this paper's own domain is 3D-LLM~\cite{hong2023threedllm}, which
injects 3D into a language model through a trained 3D feature extractor, an added
localization mechanism, and over 300k generated 3D-language examples. Decomposition
has the contrasting property: because the task leaves the model, a stock
checkpoint suffices, with no task-specific training and no fine-tuning
(Section~\ref{sec:measured}).

The alternative is to \emph{decompose}: to observe that the task is not uniformly
fuzzy, but a mixture of parts with a \emph{right answer} (parsing, lookup,
geometry, constraint satisfaction) and parts that are genuinely ambiguous or
generative. Decomposition extracts the deterministic parts to a host that solves
them exactly, and leaves the model only the residue.

Our contribution is not the idea of decomposition, which is old
~\cite{simon1962architecture} and now common in prompting~\cite{wei2022cot,zhou2023leasttomost,khot2023decomposed} (Section~\ref{sec:related}). It is threefold:
\begin{enumerate}
  \item A \textbf{measured property}, \emph{parity by construction}: on decomposed
  sub-tasks the model-size curve flattens, because capability has been removed
  from the critical path rather than improved. We report this with a per-task,
  per-model evaluation over nine models in one domain (Section~\ref{sec:measured}).
  \item A \textbf{method} for identifying where a task admits this cut, and for
  making the remaining residue safe (Section~\ref{sec:method}), together with the
  \textbf{construction} that makes such a cut exist: a deterministic surface and a
  closed grammar over it (Section~\ref{sec:construction}).
  \item \textbf{Cross-domain evidence} that the same design move recurs (physics
  from labels, motion retargeting, object-model editing, and a Web-CLI tool family;
  Section~\ref{sec:general}), presented honestly as illustrative case studies
  rather than as additional measured results. The Web-CLI instance shares an author
  and a project with the rest and is not an independent replication.
\end{enumerate}

The practical stakes are highest where small, on-device~\cite{abdin2024phi3}, or sovereign~\cite{webcli2026} models are
desirable: decomposition is what makes a $\sim$1\,GB 4-bit-quantized model~\cite{frantar2023gptq,dettmers2023kbit} sufficient without task-specific
training, because it shrinks the model's job until what remains fits even a small
model's reach.

\section{The principle}
\label{sec:principle}

\paragraph{Capability-bound tasks.} Call a task \emph{capability-bound} if its
accuracy rises with model size and plateaus below ceiling; better models help,
but the task caps out. A capability-bound task is a candidate for decomposition
precisely because adding capability yields diminishing returns~\cite{kaplan2020scaling}.

\paragraph{The deterministic substructure test.} Ask of a capability-bound task:
is there a deterministic algorithm for some sub-part? A selector over a labelled
graph is resolved by parsing and lookup, as CSS selectors are over a DOM~\cite{querySelector}; a collision body is generated from a
class label; a bone correspondence is computed from geometry. Where such an
algorithm exists, the capability it supplies is \emph{free}: a host computes it
exactly, every time, at no marginal cost and with a proof of correctness. One is
not removing intelligence from the system; one is sourcing it from determinism
instead of from the model.

\paragraph{Parity by construction.} Let $x$ be a task input, $H$ the host's
deterministic resolution, $M$ a model, $R$ the constructor of the residual task,
and $E$ the validator and executor. For a sub-task the host fully resolves,
\begin{equation}
y = H(x),
\end{equation}
which takes no model argument. Any two models therefore produce the same output on
that sub-task: not as a measured coincidence but as a structural guarantee, since
the host answered before the model was consulted. We call this
\emph{parity by construction}. For the complete system,
\begin{equation}
z = E\big(H(x),\, M(R(x,H(x)))\big),
\end{equation}
and $z$ depends on $M$ through the residual term. Parity holds for $z$ only where
$M$'s contribution is absent or fixed. Agreement across models on the residual is
an empirical question, not a consequence of the construction. The open question is
therefore not whether $H$ is model-independent, which follows by construction, but
how much of $z$'s reliability can be moved into $H$, and what building and
maintaining $H$ costs.

The guarantee rests on four stated assumptions, none of which we prove:
(A1) the input representation is valid, so $x$ carries the information $H$ needs;
(A2) $H$ is correctly implemented for the function it is meant to compute;
(A3) the host context is complete, so $R$ does not need deterministic information
that $H$ omits; and (A4) $E$ applies its verification rules correctly. Under these,
the host-resolved component is correct with respect to $H$. The end-to-end system
is not formally verified, and the measurements in Section~\ref{sec:measured} test
it only on our fixtures.

Its signature is a flat model-size curve on the decomposed sub-tasks and a sloped
curve only on the irreducible residue.

\subsection{The construction behind the cut}
\label{sec:construction}

\begin{figure}[t]\centering
\includegraphics[width=0.85\linewidth]{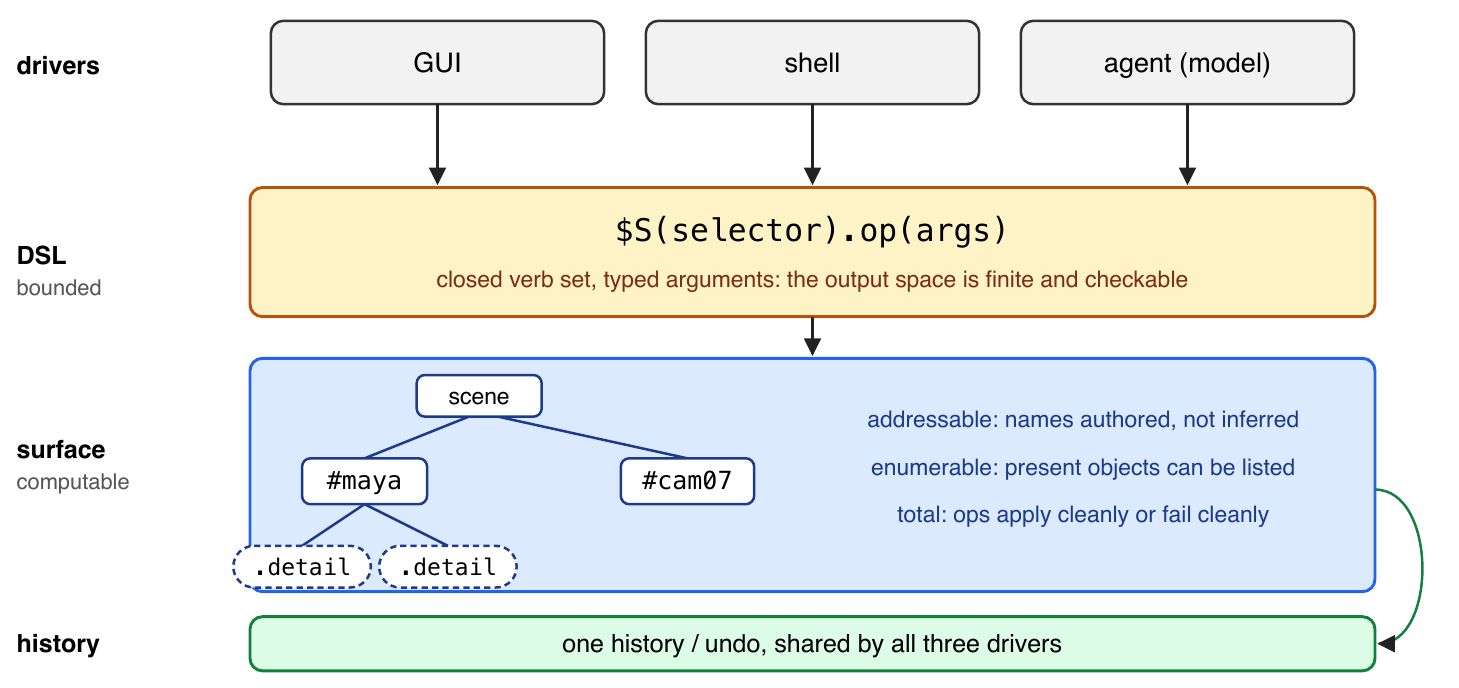}
\caption{The construction. A deterministic surface (bottom; a small labelled object
graph) makes answers computable. A DSL over it (middle) makes the model's output
bounded. The three drivers (top) enter through the same DSL; they share one model of
the scene and one history/undo, which runs beneath the surface.}
\label{fig:construction}
\end{figure}

The test above says where to look for a deterministic part; this subsection says
what has to be built for there to be one. Two halves are needed, and they are
duals (Figure~\ref{fig:construction}). A \emph{deterministic surface} makes an
answer \emph{computable}: a host can produce it exactly. A \emph{DSL}, a closed
grammar of operations over the surface, makes an answer \emph{bounded}: the
model's output space is finite and checkable. Neither works alone. In
Section~\ref{sec:measured}, selector resolution could move host-side only because
the graph was addressable, and operation selection could become a lookup only
because the verb set was closed. Without the DSL there is no finite thing to
resolve \emph{into}; without the surface there is nothing to resolve
\emph{against}. The cut the method looks for is the seam where the two meet. This is
also why a stock model suffices: given a deterministic surface and a closed grammar,
the model is never asked the capability-bound question, so there is no
task-specific capability left to train into it.

\paragraph{Three testable properties of a surface.} A domain has a deterministic
surface if it passes three tests:
\begin{itemize}
  \item \textbf{Addressable.} Every object has a stable name, \emph{assigned by an
  author}, not inferred.
  \item \textbf{Enumerable.} The set of present objects can be listed, so ``which
  parts?'' has an exhaustive answer.
  \item \textbf{Total.} Every operation either applies cleanly or fails cleanly;
  there are no partial or ambiguous states. This is what makes verification and
  undo possible.
\end{itemize}
The DSL half asks for a finite verb set with typed arguments. A DOM with ids, a
filesystem, and a relational schema pass; a photograph does not, since it has no
names to address, no list of its objects, and no operation that fails cleanly.

\paragraph{Authored labels: the precondition and the cost.} The first property
hides the method's real price: the labels must be authored. The system works
because someone named \texttt{\#maya} and \texttt{.detail}; where labels must be
inferred, the task is fuzzy again. Authoring and maintaining them is part of the
cost of the host (Section~\ref{sec:limits}).

\paragraph{Single-surface parity: a third ingredient, scoped.} A third
ingredient is that the GUI, a shell and an agent drive the \emph{same} model
through the \emph{same} operations with \emph{one} undo history (not to be confused
with parity by construction above, which concerns model size). Its necessity depends
on the claim. For \emph{LLM-drivability} it is optional: a surface plus a DSL is
enough. For an \emph{authorship} claim, that the model supplies speed while the
human retains authorship, it is load-bearing, since it holds only if a person can
take over mid-stream, inspect what the model did, and undo it in the same history;
an automation API bolted beside a GUI buys drivability and forfeits this. We make
the authorship claim as a design argument; we have not measured it.

\section{Measured result: 3D scene editing}
\label{sec:measured}

\begin{figure}[tbp]\centering
\includegraphics[width=0.8\linewidth]{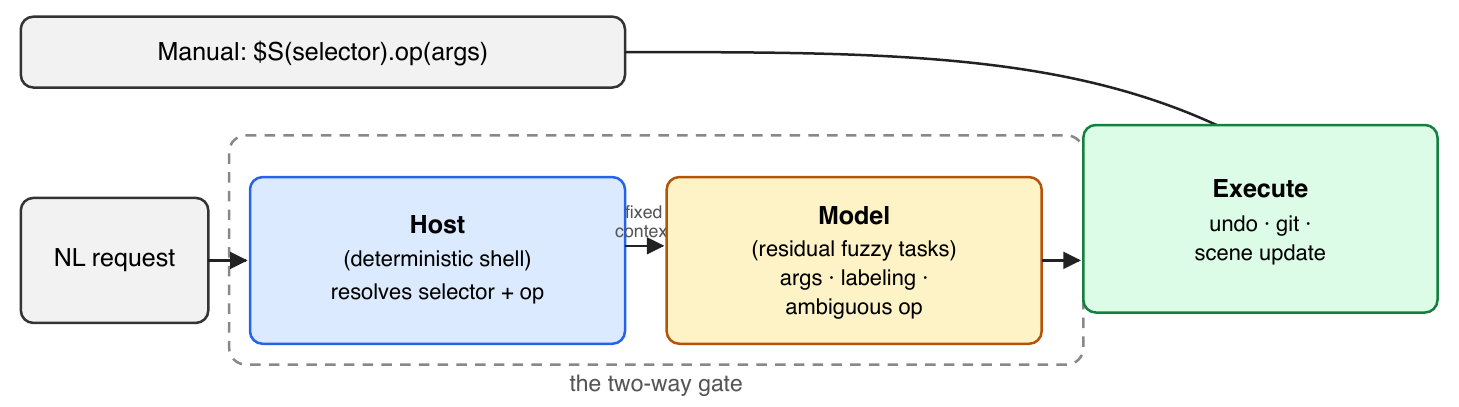}
\caption{The two-way gate. A natural-language request passes through a deterministic host, which resolves the selector and operation and hands them to the model as fixed context. The model supplies the residual arguments and ambiguous choices, and the result is executed as an undoable scene update. A manual path, \texttt{\$S(selector).op(args)}, bypasses the model entirely.}
\label{fig:gate}
\end{figure}

We evaluate in a 3D scene-editing system~\cite{strata2026} in which natural-language edits are
mapped onto a closed set of structured operations over a labelled scene graph (Figure~\ref{fig:gate}).
The task decomposes into selector resolution (which parts), operation selection
(which verb), argument extraction (which values), multi-operation segmentation,
and labelling (naming an unlabelled part). We score at the resolved-node level
over 88 fixtures: 22 each for op-selection, selector-resolution and
arg-extraction, 13 for multi-op and 9 for labelling.

\paragraph{Models.} Nine stock checkpoints, none fine-tuned or otherwise trained
for this task. Five local Qwen2.5-Coder-Instruct checkpoints~\cite{qwen25coder} at
0.5B, 1.5B, 3B, 7B and 14B parameters, served by Ollama~\cite{ollama} at Q4\_K\_M
quantization with \texttt{temperature=0}; and four frontier models accessed through
the Anthropic API: \texttt{claude-haiku-5-5}, \texttt{claude-sonnet-5-5},
\texttt{claude-opus-5-5} and \texttt{claude-fable-5-1}. The local family alone spans
a thirty-fold parameter range (494M to 14.8B).

\paragraph{Before decomposition (scaffolded baseline).} With the model asked to
compose the whole task, selector resolution is the weakest task at six of the nine
model sizes and second-weakest at the remaining three, behind multi-op
(Table~\ref{tab:scaffolded}); the two tasks that trade places are exactly the two
the host later takes over. The model-size curve is neither flat nor
clean: it runs 5\%, 55\%, 36\%, 41\%, 45\% across the local family, so the 3B,
7B and 14B checkpoints all score \emph{below} the 1.5B, and the best frontier
model reaches only 68\%. The task caps out well below ceiling for every model we
measured, which is what makes it capability-bound in the sense of
Section~\ref{sec:principle}, and a candidate for decomposition. The bare condition
(scene only, no selectors or vocabulary; overall scores in Table~\ref{tab:ladder},
per-task in the released data) makes the same
point more sharply: selector resolution scores 42\% for the 1.5B local checkpoint
against 9\%, 18\% and 18\% for Haiku, Sonnet and Fable, so the small model
outscores each frontier model by a factor of 2.3 to 4.7. We report this without a
mechanism, and each frontier figure is a single run.

\begin{table}[h]\centering
\caption{Scaffolded baseline, before host resolution: the model composes the whole
task. Local checkpoints at Q4\_K\_M; frontier models through the Anthropic API.
Where more than one repeat exists the mean is reported; the 1.5B column is the mean
of three repeats and the spread is at most 5 points on any task.}
\label{tab:scaffolded}
\footnotesize
\begin{tabular}{lrrrrrrrrr}
\toprule
task & 0.5B & 1.5B & 3B & 7B & 14B & Haiku & Sonnet & Opus & Fable\\
\midrule
op-selection        & 9\%  & 76\% & 77\% & 95\% & 95\% & 95\%  & 100\% & 95\%  & 100\%\\
selector-resolution & 5\%  & 55\% & 36\% & 41\% & 45\% & 59\%  & 59\%  & 50\%  & 68\%\\
arg-extraction      & 68\% & 91\% & 86\% & 95\% & 95\% & 95\%  & 95\%  & 95\%  & 95\%\\
multi-op            & 0\%  & 54\% & 46\% & 38\% & 62\% & 77\%  & 77\%  & 69\%  & 85\%\\
labelling           & 67\% & 78\% & 67\% & 78\% & 89\% & 89\%  & 89\%  & 100\% & 100\%\\
\midrule
overall             & 27\% & 71\% & 64\% & 72\% & 77\% & 83\%  & 84\%  & 81\%  & 89\%\\
\bottomrule
\end{tabular}
\end{table}

\begin{figure}[tbp]\centering
\includegraphics[width=0.8\linewidth]{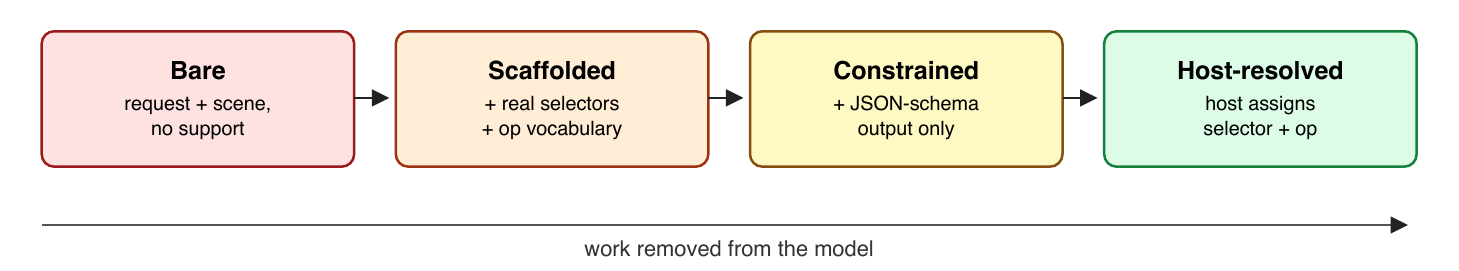}
\caption{Work removed from the model across the harness ladder: a bare request with the scene, a scaffolded prompt with real selectors and the op vocabulary, constrained JSON-schema output, and host-resolved selector and op. All four rungs send a prompt of comparable size (10--11.5k tokens for the local tokenizer); the rungs differ in what the prompt \emph{contains}, not in how much of it there is.}
\label{fig:ladder}
\end{figure}

\begin{table}[h]\centering
\caption{Overall score at each rung of the harness ladder (Figure~\ref{fig:ladder}),
for the four models measured at all four rungs. The ladder is not monotonic:
constrained output scores \emph{below} scaffolded for every one of these models,
and improves only on bare. Host resolution is the only rung that produces a flat
curve across models.}
\label{tab:ladder}
\footnotesize
\begin{tabular}{lrrrr}
\toprule
rung & 1.5B & Haiku & Sonnet & Fable\\
\midrule
bare           & 55\% & 43\% & 53\% & 57\%\\
scaffolded     & 71\% & 83\% & 84\% & 89\%\\
constrained    & 70\% & 81\% & 77\% & 83\%\\
host-resolved  & 88\% & 86\% & 89\% & 90\%\\
\bottomrule
\end{tabular}
\end{table}

\paragraph{After decomposition.} Moving selector resolution and unambiguous
operation selection host-side, and splitting multi-operation requests where the
split is unambiguous, resolves them from the request text before the model is
consulted; ambiguous cases are deferred to the model. This removes them from the
model's job (Figure~\ref{fig:ladder} traces how the model's share shrinks).
Selector resolution, unambiguous op-selection and unambiguous multi-op segments run
host-side; argument extraction, ambiguous choices and labelling stay with the model.
Parity by construction applies to the host-assigned slots. The instrumentation
records the same counters in \emph{every} host-resolved run, for all nine models:
the host assigned the selector in 31 of 32 emitted operations and the operation in
31 of 32, deferring one genuinely ambiguous case. The arguments, which the model
still produces, are not covered by the construction; their agreement is a measured
result.

\begin{table}[h]\centering
\caption{After host resolution. Scores on each task, scored on resolved nodes.
Labelling ($\dagger$) is not decomposed; it is the residual the model still owns.
Op-selection is identical for all nine models and selector-resolution for eight of
nine.}
\label{tab:resolved}
\footnotesize
\begin{tabular}{lrrrrrrrrr}
\toprule
task & 0.5B & 1.5B & 3B & 7B & 14B & Haiku & Sonnet & Opus & Fable\\
\midrule
op-selection        & 86\% & 86\% & 86\% & 86\% & 86\% & 86\% & 86\% & 86\%  & 86\%\\
selector-resolution & 91\% & 91\% & 91\% & 91\% & 91\% & 86\% & 91\% & 91\%  & 91\%\\
arg-extraction      & 86\% & 91\% & 91\% & 91\% & 91\% & 95\% & 91\% & 91\%  & 91\%\\
multi-op            & 85\% & 85\% & 85\% & 85\% & 85\% & 77\% & 85\% & 77\%  & 85\%\\
labelling$^\dagger$ & 67\% & 78\% & 67\% & 78\% & 89\% & 78\% & 89\% & 100\% & 100\%\\
\midrule
overall             & 85\% & 88\% & 86\% & 88\% & 89\% & 86\% & 89\% & 89\%  & 90\%\\
\bottomrule
\end{tabular}
\end{table}

\paragraph{The flat line.} Once the host resolves the context, operation selection
scores \textbf{86\% for every one of the nine models}, from a 494M local checkpoint
to a frontier Opus-class model, and selector resolution scores 91\% for eight of the
nine. These are not scores that agree within noise; they are the same score. At the
fixture level the agreement is equally sharp: comparing each model against the
0.5B checkpoint, op-selection and selector-resolution agree on 22 of 22 fixtures for
every model but Haiku, and multi-op on 13 of 13 for every model but Haiku and Opus
(Table~\ref{tab:agreement}). Task
performance on the host-assigned slots no longer depends on which model answers.

\begin{table}[h]\centering
\caption{Fixture-level agreement with the 0.5B checkpoint under host resolution.
Each cell counts fixtures on which the two models reach the same outcome.}
\label{tab:agreement}
\footnotesize
\begin{tabular}{lrrrrr}
\toprule
model & op-sel. & selector & arg-extr. & multi-op & labelling\\
\midrule
1.5B   & 22/22 & 22/22 & 21/22 & 13/13 & 8/9\\
3B     & 22/22 & 22/22 & 21/22 & 13/13 & 7/9\\
7B     & 22/22 & 22/22 & 21/22 & 13/13 & 8/9\\
14B    & 22/22 & 22/22 & 21/22 & 13/13 & 7/9\\
Haiku  & 20/22 & 21/22 & 20/22 & 12/13 & 8/9\\
Sonnet & 22/22 & 22/22 & 21/22 & 13/13 & 7/9\\
Opus   & 22/22 & 22/22 & 21/22 & 12/13 & 6/9\\
Fable  & 22/22 & 22/22 & 21/22 & 13/13 & 6/9\\
\bottomrule
\end{tabular}
\end{table}

\paragraph{The widest single comparison.} The 0.5B checkpoint moves from 5\% to
91\% on selector resolution, 9\% to 86\% on op-selection and 0\% to 85\% on
multi-op, with no change to the model and no training of any kind: the same stock
494M checkpoint, asked a smaller question. The failure mode is as informative as
the score. Under scaffolding, 51 of its 88 outcomes are \texttt{parse\_error} --- the
model emits text containing no readable operation at all. Under host resolution it
produces zero parse errors; every emission is well-formed and 13 are simply wrong.
The small model does not fail the undecomposed task by answering badly, but by being
unable to compose a well-formed operation; once the host supplies the selector and
the verb, composition is no longer asked of it.

\paragraph{Cost, not only accuracy.} Decomposition also shortens what the model
must produce. For the 0.5B checkpoint, mean completion length falls from 1{,}385
tokens under scaffolding to 26 under host resolution, and mean per-call latency from
2{,}978\,ms to 275\,ms --- a complete run going from 92.3\,s to 11.3\,s despite
host resolution issuing \emph{more} model calls (41 against 31). Where the aim is an
on-device or sovereign deployment, the decomposed path is both more accurate and an
order of magnitude cheaper to run.

\begin{figure}[!t]\centering
\includegraphics[width=0.8\linewidth]{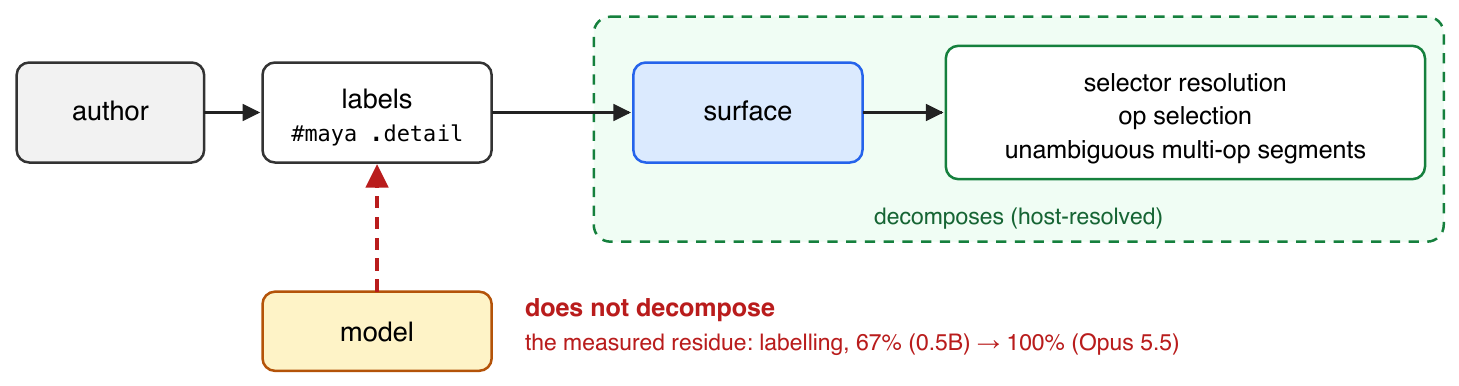}
\caption{Where the labels come from. Everything downstream of the surface is
resolved by the host. The one task that does not decompose, labelling, is the step
that would replace the author: the measured residue (Table~\ref{tab:resolved}).}
\label{fig:labels}
\end{figure}

\paragraph{The honest residue.} Decomposition has a floor. Labelling (naming an
unlabelled shape from its descriptors) does \emph{not} decompose; it is genuinely
generative, and it is the only task in Table~\ref{tab:resolved} that still moves
with model capability, running 67\%, 78\%, 67\%, 78\%, 89\% across the local family
and reaching 100\% at Opus and Fable --- sloped across exactly the range on which
the other four tasks are flat. Multi-op segmentation retains residual ambiguity on 2
of 13 fixtures. These are not failures of the method but its correct boundary: the
method extracts the deterministic parts and \emph{stops} at the irreducible core.
The boundary is also consistent with the construction of
Section~\ref{sec:construction}: labelling is the task of producing the authored
labels the surface depends on, so the method's boundary is exactly the method's
input (Figure~\ref{fig:labels}).

\paragraph{Two models that do not fit the pattern.} We report two deviations
rather than smoothing them. \texttt{claude-haiku-5-5} is the only model whose
host-resolved scores differ on three tasks (86\% on selector-resolution, 95\% on
arg-extraction, 77\% on multi-op), and the only one to disagree with the 0.5B
checkpoint on op-selection. \texttt{claude-opus-5-5} also scores 77\% on multi-op.
We do not have an account of either; both are reported in
Table~\ref{tab:resolved} and the per-fixture data is released.

\paragraph{Data availability.} All 36 analysed runs, with per-fixture outcomes,
failure kinds and per-run provenance (model digest, quantization, server version,
decode settings, scorer hash, system commit), are released as
\texttt{eval-results.jsonl}; the full tables are in \texttt{RESULTS.md}. Both files
are arXiv ancillary files, and are mirrored at
\url{https://github.com/tejaswigowda/strata-editor/tree/013417d/docs/anc} (pinned to
commit \texttt{013417d}). The call-level capture (prompts as sent, raw responses,
token usage and latency), from which the completion-length and latency figures above
are taken, is not part of these files. Two further
runs were excluded and are recorded with their reason: both reported a completed
harness loop while the majority of their API calls raised exceptions, and both have
clean replacement repeats. The system under test is \texttt{strata-editor} at
commits \texttt{888d97e} and \texttt{6b7cbb7}; the scorer is
\texttt{sha256:43698355\dots} for every run reported here.

\section{Method}
\label{sec:method}

For any task one is tempted to delegate whole to a model, apply in order:
\begin{enumerate}
  \item \textbf{Find the right-answer part.} Is there a sub-task with a
  deterministic solution (parse, lookup, geometry, constraint)? Extract it to the
  host. Section~\ref{sec:construction} gives what the domain must offer for this
  to be possible.
  \item \textbf{Isolate the residue.} What genuinely remains open? That, and only
  that, is the model's job.
  \item \textbf{Resolve the context (``give sight, not control'').} Hand the model
  the host-resolved facts as fixed context so it decides only the minimum. In the
  measured domain the host resolves the selector and shows it to the model as
  fixed context rather than asking the model to produce one.
  \item \textbf{Verify the output.} If the residue's output admits a deterministic
  check (valid operation enum; left-side bones on the left), apply it: wrong
  generative answers are caught, not shipped.
  \item \textbf{Accept the irreducible.} Where no deterministic substructure
  remains, let the model do it and let capability show. Do not force a genuinely
  generative task into a brittle deterministic rule.
\end{enumerate}

\section{Generalization}
\label{sec:general}

The same move recurs across domains. We present these as case studies of the
design discipline; \textbf{they are not separately measured parity studies}, and
we flag each as such. The first three sit inside one system; the fourth does not.

\paragraph{Physics from labels \illus.} Rigid-body setup is treated as a
deterministic lookup: class labels on scene nodes (\texttt{.static},
\texttt{.dynamic}, \texttt{.kinematic}, \texttt{.trigger}) generate collision
bodies host-side. ``Set up physics'' is thereby decomposed into a label table; no
model and no per-instance code performs it.

\paragraph{Motion retargeting with a narrow model \illus.} Retargeting motion
capture onto an arbitrary skeleton is deterministic geometry~\cite{gleicher1998retargetting} (world-delta transfer
with reference-pose alignment). A model is admitted only for the fuzzy residue
(labelling unknown bones from a closed enum), and the host verifies each label
geometrically before accepting it: decomposition followed by verification,
exactly the method of Section~\ref{sec:method}.

\paragraph{Object-model editing \illus.} A selector-and-operation layer~\cite{3dom2026} over a
scene graph keeps resolution and the operation set deterministic; a model, when
present, supplies only argument values and genuinely ambiguous choices. The
deterministic core is usable with no model at all.

\begin{figure}[!t]\centering
\includegraphics[width=0.85\linewidth]{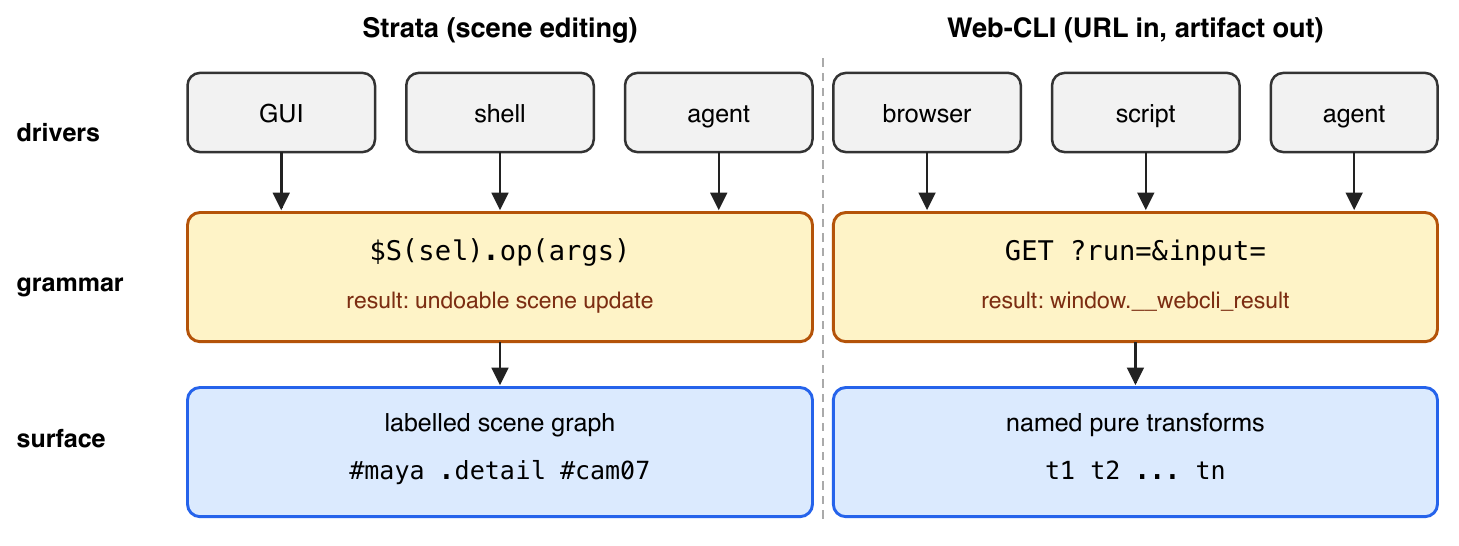}
\caption{The same construction in a second domain. Left: Strata. Right: the
Web-CLI family. Both have a deterministic surface (bottom), a closed grammar over it
(middle) and several drivers entering through one entry point (top). Same author and
same project: this is not an independent replication.}
\label{fig:second}
\end{figure}

\paragraph{A second instance: the Web-CLI tool family \illus.} The Web-CLI
family~\cite{webcli2026} was built for a different problem, \texttt{URL in,
artifact out}: tools are \texttt{GET}-only, selected by \texttt{run=} parameters
over a set of pure transforms, and report their output through a
\texttt{window.\_\_webcli\_result} handle. It is the same construction
(Figure~\ref{fig:second}): the transforms are a named, deterministic surface; the
URL parameters are a closed grammar over it; and a person in a browser, a script
and an agent all use the same entry point. The caveat belongs in the text: this
is the same author in the same project, so it is \emph{not} an independent
replication. The honest claim is that the construction \emph{recurred under the same
constraint}, which is suggestive of generality, stronger than an argument and
weaker than a replication.

Across these, the invariant is: the deterministic substructure is extracted and
made correct by construction; the model, if present, is confined to the residue
and its output is verified. What varies is only the domain.

\section{Related work}
\label{sec:related}

Our position sits within \textbf{neuro-symbolic AI}~\cite{garcez2023neurosymbolic}:
the combination of symbolic and neural components is long-standing. We claim neither a
new architecture nor the idea of combination; we contribute a measured property
(parity by construction) and a method for interactive systems.
\textbf{Tool-augmented language models} (surveyed in~\cite{mialon2023augmented}; program-aided LMs~\cite{gao2023pal,chen2023pot},
Toolformer~\cite{schick2023toolformer}, ReAct~\cite{yao2023react}; visual programming~\cite{gupta2023visprog}, HuggingGPT~\cite{shen2023hugginggpt}, and Chameleon~\cite{lu2023chameleon}) expand the model with
callable tools but leave the model orchestrating and deciding; our emphasis is to
move the capability-bound \emph{decision itself} off the model (the host resolves;
the model is never asked), not merely to equip the model with a tool.
\textbf{Generate-and-verify / LLM-Modulo} approaches~\cite{kambhampati2024llmmodulo}
share our verification step; we add decomposition \emph{before} generation: the
deterministic parts are resolved so the model never generates them, and the
measured model-independence that follows. \textbf{Constrained decoding and
grammars}~\cite{geng2023gcd,scholak2021picard,willard2023efficient,yin2017syntactic} are a tool we use, not the thesis; we measure it as one rung of the ladder (Figure~\ref{fig:ladder}, Table~\ref{tab:ladder}) and find it improves on bare prompting but not on scaffolding, and does not produce the flat curve that host resolution does. \textbf{Retrieval and
grounding}~\cite{lewis2020rag} resolve context for the model, as our step~3 does, but do
not characterize the resulting capability-independence.

\textbf{Decomposition in prompting.} Least-to-most prompting~\cite{zhou2023leasttomost} and decomposed prompting~\cite{khot2023decomposed} have the model produce a decomposition and solve the sub-problems in turn, and chain-of-thought prompting~\cite{wei2022cot} elicits intermediate steps from the model. These methods still route the decomposition itself through the model. Distillation~\cite{hsieh2023distilling} transfers such capability from a large teacher into a small student; we remove the deterministic sub-parts from the model's job instead, so there is no capability to transfer.

\textbf{Grounding references in 3D scenes.} Learned grounding of natural-language references to 3D objects~\cite{chen2020scanrefer,achlioptas2020referit3d} addresses the problem that selector resolution addresses. Where the scene graph carries labels, the same resolution can be done deterministically host-side.

\textbf{Injecting 3D into language models.} 3D-LLM~\cite{hong2023threedllm} takes 3D
point clouds in and produces language out (captioning, question answering,
grounding, dialogue, navigation). It does so by training a 3D feature extractor
over multi-view renders, adding a localization mechanism, and training on over 300k
generated 3D-language examples: the expansion path in this paper's own domain,
with its cost made explicit as a trained pipeline and a generated corpus. We do not
compare the two. They share no benchmark, task or output, and we
report no numbers across them. They are complementary. 3D-LLM's problem is
grounding language in \emph{unlabelled} 3D, which is the regime our first surface
property excludes (addressable: names assigned by an author, not inferred) and
where Section~\ref{sec:limits} says the construction does not reach; on an
unlabelled point cloud there is nothing to address, enumerate or operate on. It is
also the difficulty that the one task we measure as refusing to decompose,
labelling, isolates. The two approaches meet the same difficulty from opposite
sides: one trains a model to produce the labels; the other requires that an author
supply them, and measures what that buys. The method's identified boundary is the
other work's problem statement.

The distinction we draw is one of emphasis with measured support: expansion makes
correctness a function of capability; decomposition makes the decomposed part
independent of it.

\section{Limitations}
\label{sec:limits}

Our measured evidence is one domain and a small fixture set (9 to 22 fixtures per
task); the cross-domain claims are illustrative, not measured, and the Web-CLI
instance is by the same author in the same project, so the two instances are
suggestive rather than independent. A single additional \emph{measured} domain would
move the central claim from ``demonstrated once'' to ``demonstrated across domains''
and is the most valuable next step.

Three properties of the measurement itself should be read with the numbers.
\emph{First}, \texttt{temperature=0} is forwarded for the local models but is
\textbf{not} forwarded on the Anthropic client path for Claude 5 models, so the
frontier columns are sampled rather than greedy; this is recorded per run in the
released provenance. \emph{Second}, repeats exist for the 0.5B and 1.5B checkpoints
and for two frontier conditions, but most frontier cells are a single run; the local
repeats vary by at most 5 points on any task, and the host-resolved local cells do
not vary at all. \emph{Third}, the sweep spans two commits of the system under test
(\texttt{888d97e} and \texttt{6b7cbb7}); the scorer hash is identical across all
runs, so the scoring is comparable, but the harness is not byte-identical across
the whole sweep.

The method's hardest judgment, correctly distinguishing deterministic from fuzzy
substructure, can fail: a deterministic rule applied where the world is actually
variable is brittle-by-construction (a bone-name table assumes predictable naming
and fails on the rig that does not), which is why steps~4--5 retain a verified fuzzy
fallback. Decomposition is not free: it moves work from the model's runtime to a
host that must be built and maintained, and the price includes the labels
themselves, which must be authored (Section~\ref{sec:construction}); the absence of
task-specific training is a cost moved, not removed. This cost and the floor are the
same fact: the one measured task that refuses to decompose is labelling, so the
method's boundary is exactly the method's input, and the generative residue it
leaves remains capability-bound.

\paragraph{Scope.} The construction applies to domains with discrete, nameable
objects and a finite set of operations over them: scene graphs, the DOM,
filesystems, relational schemas, infrastructure manifests, version control. It
does not obviously reach continuous, unlabelled domains such as image synthesis,
prose voice or musical phrasing, and we do not claim that it does. Nor does it reach
unlabelled 3D, the setting of 3D-LLM (Section~\ref{sec:related}): the labels are
the input, and producing them is the residue. The authorship
argument for single-surface parity is a design argument, not a measured result.

\section{Conclusion}

When a model is not good enough for a task, the reflex is to reach for a better
model. Often the better move is to reach for a smaller task. Decomposition
extracts the deterministic substructure, makes it correct by construction, and
confines the model to the irreducible residue. Across nine models spanning a
494M-parameter local checkpoint to a frontier Opus-class model, the decomposed
sub-tasks score identically while the one task that does not decompose stays sloped
over the same range --- so that a stock checkpoint suffices, with no task-specific
training or fine-tuning, and runs an order of magnitude cheaper. The principle is
old~\cite{simon1962architecture}; the measured property and the method are the
contribution, and the cross-domain recurrence suggests it is a general discipline
for building reliable, verifiable, and sovereign AI systems rather than a trick of
one domain.

\bibliographystyle{plain}
{\footnotesize
\bibliography{refs}
}

\end{document}